\documentclass[10pt,twocolumn,letterpaper]{article}

\usepackage{wacv}
\usepackage{times}
\usepackage{epsfig}
\usepackage{graphicx}
\usepackage{amsmath}
\usepackage{amssymb}
\usepackage{verbatim}
\usepackage{mathtools}
\usepackage{algorithm}% http://ctan.org/pkg/algorithm
\usepackage{algpseudocode}% http://ctan.org/pkg/algorithmicx
\usepackage{stfloats}

\usepackage{verbatim}

\def\wacvPaperID{1223} % Enter the WACV Paper ID here

\wacvfinalcopy % *** Uncomment this line for the final submission

\ifwacvfinal
\def\assignedStartPage{9876} % *** Enter the assigned starting page number (instead of 9876)
\fi

\ifwacvfinal
\usepackage[breaklinks=true,bookmarks=false]{hyperref}
\else
\fi

\ifwacvfinal
\else
\fi

\begin{document}

%%%%%%%%% TITLE

% ArXiv: Adaptive WGAN with loss change rate balancing 
\title{Accelerated WGAN update strategy with loss change rate balancing}
%\title{\textcolor{red}{Accelerated adaptive WGAN update strategy with loss change rate balancing}}

\author{Xu Ouyang\\
Illinois Institute of Technology\\
\and
Gady Agam\\
Illinois Institute of Technology\\
}

\maketitle
%\thispagestyle{empty}

%%%%%%%%% ABSTRACT
\begin{abstract}
   Optimizing the discriminator in Generative Adversarial Networks (GANs) to completion   
   in the inner training loop is computationally prohibitive, and on finite datasets 
   would result in overfitting. To address this, a common update strategy is to alternate 
   between k optimization steps for the discriminator D and one optimization step for the generator G. 
   This strategy is repeated in various GAN algorithms where k is selected empirically. In this paper, 
   we show that this update strategy is not optimal in terms of accuracy and convergence speed,
   and propose a new update strategy for networks with Wasserstein GAN (WGAN) group related loss functions 
   (\eg WGAN, WGAN-GP, Deblur GAN, and Super resolution GAN). The proposed update strategy is based on a loss 
   change ratio comparison of G and D. We demonstrate that the proposed strategy improves
   both convergence speed and accuracy.
\end{abstract}

%%%%%%%%% BODY TEXT
%%%%%%%%% Introduction
\section{Introduction}
GANs~\cite{NIPS2014Goodfellow} provide an effective deep neural network framework that can capture data distribution. GANs are modeled as a min-max two-player game between a discriminator network $D_\psi(x)$ and 
a generator network $G_\theta(z)$. The optimization problem solved by GAN~\cite{NIPS2017Nagarajan} is given by:
\begin{multline}
\mathop{min}_{G}\mathop{max}_{D}V(G,D)=\mathbb{E}_{x\backsim p_{data}}[f(D(x))]+\\
\mathbb{E}_{z\backsim p_{latent}}[f(-D(G(z)))]
\label{eq: objective function}
\end{multline}
where $G:Z \rightarrow X$ maps from the latent space Z to the input space X; $D:X \rightarrow \mathbb{R}$ maps from the input space to a classification of the example as fake or real; and $f:\mathbb{R} \rightarrow \mathbb{R}$ is a concave function. 
In the remainder of this paper, we use the Wasserstein GAN~\cite{Arxiv2017Arjovsky} obtained when using $f(x) = x$.

\begin{figure}[th]
\centerline{
\begin{tabular}{c}
  \resizebox{0.5\textwidth}{!}{\rotatebox{0}{
  \includegraphics{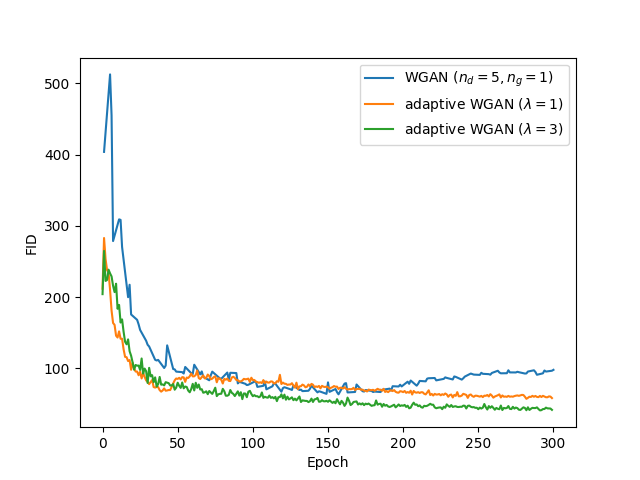}}}
\end{tabular}}
\caption{Comparison of the Frenchet Inception Distance (FID) for WGAN and the proposed adaptive WGAN with different  coefficient $\lambda$ on the CIFAR10 dataset. A lower FID means better performance. The parameters $n_d$ and $n_g$ show the fixed number of update steps in WGAN for the discriminator and generator respectively.}
\label{fig: WGAN training curve}
\end{figure} 

GANs have been shown to perform well in various image generation applications such as: deblurring images~\cite{CVPR2018Kupyn}, increasing the resolution of images~\cite{CVPR2017Ledig}, generating captions from images~\cite{AAAI2019Chen}, and generating images from captions~\cite{ICPR2018Xu}. 
Training GANs may be difficult due to stability and convergence issues. To understand this consider the fact that GANs minimize a probabilistic divergence between real and fake (generated by the generator) data distributions~\cite{NIPS2016Nowozin}. Arjovsky \etal~\cite{ICLR2017Arjovsky} showed that this divergence may be discontinuous with respect to the parameters of the generator, and may have infinite values if the real data distribution and the fake data distribution do not match. 

In order to solve a divergence continuous problem, WGAN~\cite{Arxiv2017Arjovsky} uses the Wasserstein-divergence by removing the sigmoid function in the last layer of the discriminator 
and so restricting the discriminator to Lipschitz continuous functions instead of the Jensen-Shannon divergence in the original GAN~\cite{NIPS2014Goodfellow}. WGAN will always converge when the
discriminator is trained until convergence. However, in practice, WGAN is trained with a fixed number 
(five) of discriminator update steps for each generator update step.

Even though WGAN is more stable than the original GAN, Mescheder \etal~\cite{ICML2018Mescheder} proved that WGAN trained with simultaneous or alternating gradient descent steps with a fixed number of discriminator updates per generator update and a fixed learning rate $\alpha > 0$ does not converge to the Nash equilibrium for a Dirac-GAN, where the Dirac-GAN~\cite{ICML2018Mescheder} is a simple prototypical GAN.

The WGAN update strategy proposed in~\cite{Arxiv2017Arjovsky} suggests an empirical ratio of five update steps for the discriminator to one update step for the generator. As should be evident this empirical ratio should not hold in all cases. Further, this ratio need not be fixed throughout training. Maintaining an arbitrary fixed ratio is inefficient as it will inevitably lead to unnecessary update steps. 
In this paper, we address the selection of the number of training steps for the discriminator and generator and show how to adaptively change them during training so as to make the training converge faster to a more accurate solution. We focus in this work on WGAN related loss as it is more stable than the original GAN. 

The strategy we propose for balancing the training of the generator and discriminator is based on 
the discriminator and generator loss change ratios ($r_d$ and $r_g$, respectively). 
Instead of a fixed update strategy we decide whether to update the Generator or discriminator by comparing the weighted loss change ratios $r_d$ and $\lambda \cdot r_g$ where the weight $\lambda$ is a hyper-parameter assigning coefficient to $r_g$. 
The default value $\lambda=1$ leads to higher convergence speed and better performance in nearly all the models we tested and is already superior to a fixed strategy. It is possible to further optimize this parameter for additional gains either using prior knowledge (\eg giving preference to training the generator as in the original WGAN) or in empirical manner. Note, however, that the proposed approach accelerates convergence even without optimizing $\lambda$ .

In addition to the acceleration aspect of the proposed update strategy, we studied its convergence properties. To do so we followed the methodology by Mescheder \etal~\cite{ICML2018Mescheder}. Following this methodology, we demonstrate that the proposed strategy can reach a local minimum point for the Dirac-GAN problem, whereas the original update strategy cannot achieve it. 

To further demonstrate the advantage of our strategy, we train the WGAN~\cite{Arxiv2017Arjovsky}, WGAN-GP~\cite{NIPS2017Gulrajani}, Deblur-GAN~\cite{CVPR2018Kupyn}, and SR-WGAN~\cite{CVPR2017Ledig} using the proposed strategy using different image datasets. These represent a wide range of WGAN applications. 
Experimental results show that in general the proposed strategy converges faster while achieving in many cases better accuracy. An illustration is provided in Figure~\ref{fig: WGAN training curve} where the proposed adaptive WGAN is compared with the traditional WGAN update strategy. 

The main contribution of this paper is proposing an adaptive update strategy for WGAN instead of the traditional fixed update strategy in which the update rate is set empirically. This results in accelerated training and in many cases with higher performance. Following a common convergence analysis procedure, we show that the proposed strategy can reach local convergence for Dirac-GAN, unlike the traditional fixed update strategy which cannot do so. Experimental results on several WGAN problems with several datasets show that the proposed adaptive update strategy results in faster convergence and higher performance.

%%%%%%%%% Related work
\section{Related work}
\label{sec:related-work}
The question of which training methods for GANs actually converge was inversigated by
Mescheder \etal~\cite{ICML2018Mescheder} where they introduce the Dirac-GAN. The Dirac-GAN
consists of a generator distribution $p_\theta = \delta_\theta$ and a linear discriminator $D_{\psi}(x)=\psi \cdot x$. 
In their paper they prove that a fixed point iteration $F(x)$ is locally convergent to $x$, when the absolute values of the eigenvalues of the Jacobian $F^\prime(x)$ are all smaller than 1. They further prove that for the Dirac-GAN, with both simultaneous and alternative gradient descent updates, the absolute values of the eigenvalues of the Jacobian $F^\prime(x)$ in GANs with unregularized gradient descent (which include the original GAN, WGAN and WGAN-GP) are all larger or equal to 1, thus showing that these types of GANs are not necessarily locally convergent for the Dirac-GAN. 
To address this convergence issue, Mescheder \etal~\cite{ICML2018Mescheder} added gradient penalties to the GAN loss and proved that regularized GAN with these gradient penalties can reach local convergence. This solution does not apply to WGAN and WGAN-GP which remain not locally convergent problems. Note that the WGAN is generally more stable and easier to train compared with GAN and hence the need for the adaptive update scheme we propose in this paper.

Heusel \etal~\cite{NIPS2017Martin} attempt to address the convergence problem in a different way by altering the learning rate. In their approach they use a two time-scale update rule (TTUR) for training GANs with stochastic gradient descent using arbitrary GAN loss functions. Instead of empirically setting the same learning rate for both the generator and discriminator, TTUR uses different learning rates for them. This is done in order to address the problem of slow learning for regularized discriminators. They prove that training GANs with TTUR can converge to a stationary local Nash equilibrium under mild condition based on stochastic approximation theory. In their experiments on image generation, the show that WGAN-GP with TTUR gets better performance. Note however that empirically setting the learning rate is generally difficult and even more so when having to set two learning rates jointly. This makes applying this solution more difficult.

It is well understood that the complexity of the generator should be higher than that of the discriminator, a fact that makes GANs harder to train. Balancing the learning speed of the generator and discriminator is a fundamental problem. Unbalanced GANs~\cite{Arxiv2020Ham} attempt to address this by pre-training the generator using variational autoencoder (VAE~\cite{ICLR2014Kingma}), and using this pre-trained generator to initialize the GAN weights during GAN training. An alternative solution is proposed in BE-GAN~\cite{Arxiv2017Berthelot}, where the authors introduce an equilibrium hyper-parameter ($\mathbb{E}[f(-D(G(z)))]/\mathbb{E}[f(D(x)]$) to maintain the balance between the generator and discriminator. 
%% NOT SO CLEAR
Training the two neural networks in this approach is time consuming and the equilibrium hyper-parameter is not suitable for all GAN training cases. For example it is not suitable when there is a content loss in the generator loss term $L_G$ as $L_G$ and the discriminator loss $L_D$ are not on the same scale. 

A similar issue to the unbalanced training of the generator and discriminator in GANs arises in imbalanced training of multiple task networks. The GradNorm~\cite{ICML2018Chen} approach provides a solution to balancing multitask network training based on gradient magnitudes. In this approach, the authors multiply each of the single-task loss terms by weights, and automatically update those weights by computing a gradient normalization term. A relative inverse training rate ($L_{\mbox{current}}/L_{\mbox{initial}}$) for each task is used to compute this normalization term. This strategy depends on a common loss term minimization where individual task terms are weighted and so is not suitable for GANs where there is no shared layer as in multi-task networks.

%%%%%%%%% Method
\section{Method}
%%%%%%%%% adaptive WGAN
\subsection{Adaptive update strategy}
In this section, we present our proposed update strategy to automatically set the update rate of the generator and discriminator instead of using a fixed rate as is commonly done.
In WGAN or any GAN based on the related WGAN loss, the Nash equilibrium is reached when the generator and discriminator loss terms stop changing. That is: 
\begin{equation}
%\begin{split}
| L_g^c-L_g^p | =0 ~~~~\&~~~~ 
| L_d^c-L_d^p | =0
%\end{split}
\label{eq: loss equal}
\end{equation}
where $L_g^c$, $L_d^c$ represent the generator and discriminator loss in the current iteration respectively, and $L_g^p$, $L_d^p$ represent the respective loss terms in the previous iteration.
Since we play a min-max game in WGAN, it is crucial to balance the generator and discriminator losses. 
The loss terms $L_g^c$, $L_d^c$ are given by:
\begin{equation}
\begin{split}
L_g^c &= E_{p(z)}[f(D_\psi(G_\theta(z)))]\\
L_d^c &= E_{ p_D(x)}[f(D_\psi(x))]+E_{p(z)}[f(-D_\psi(G_\theta(z)))]
\end{split}
\label{eq: g and d loss}
\end{equation}
Comparing $L_g^c$, $L_d^c$ directly to decide on an update policy is not possible because they are on different scales and so we define relative loss terms that can be compared. The relative loss terms are defined by computing the difference between the current and previous loss values and normalizing the difference by the loss magnitude. The relative change loss terms for the generator and discriminator are defined by:
\begin{equation}
\begin{split}
r_g=|(L_g^c-L_g^p)/L_g^p|\\
r_d=|(L_d^c-L_d^p)/L_d^p|
\end{split}
\label{eq: rg and rd}
\end{equation}

To prioritize the update of one component over the other as commonly done in GANs, we use an coefficient $\lambda$. Thus, if $r_d > \lambda \cdot r_g$, we update the discriminator, or otherwise update the generator. A larger loss change ratio of one component means that this component is in greater need for update. The details of our proposed adaptive WGAN are provided in Algorithm~\ref{alg: WGAN with our updating strategy}.

\begin{algorithm}
  \caption{Proposed adaptive update strategy}
  \label{alg: WGAN with our updating strategy}
  \begin{itemize}
    \setlength\itemsep{0.002em}
	\item {\bf parameters}: learning rate ($\alpha$); clipping parameter ($c$); loss coefficient ($\lambda$); batch size ($m$).
	\item {\bf variables}: generator parameters ($\theta$); discriminator parameters ($\psi$); generator loss change ratio ($r_g$); discriminator loss change ratio ($r_d$). The loss change ratios are initialized to 1. 
  \end{itemize}
  \begin{algorithmic}[1]
   \While{{$\theta$ has not converged}}
  	\State 
	Sample ${\lbrace x^{(i)} \rbrace}_{i=1}^m \backsim \mathbb{P}_r$ a batch from the real data    
	\State 	
	Sample ${\lbrace z^{(i)} \rbrace}_{i=1}^m \backsim p(z)$ a batch of prior samples
    	   \If{$r_d>\lambda \cdot r_g$} \qquad \# update the discriminator
           \State 
           $g_\psi \leftarrow \nabla_\psi[\frac{1}{m}\sum_{i=1}^{m}f_\psi(x^{(i)})-$
           \State
           $~~~~~~~~~~~~~~~~~~\frac{1}{m}\sum_{i=1}^{m}f_\psi (g_\theta (z^{(i)}))]$
           \State
           $\psi \leftarrow \psi+\alpha \cdot \mbox{RMSProp}(\psi,g_\psi)$
           \State
           $\psi \leftarrow clip(\psi, -c, c)$ 
          \Else   \qquad \qquad \qquad \qquad \# update the generator
          \State
          $g_\theta \leftarrow -\nabla_\theta \frac{1}{m}\sum_{i=1}^{m}f_\psi (g_\theta (z^{(i)}))$
	    \State
           $\theta \leftarrow \theta-\alpha \cdot \mbox{RMSProp}(\theta,g_\theta)$
	\EndIf
    \If {first iteration}
    	\State
	$L_g^p,L_d^p=L_g,L_d$
    \EndIf
	\State
	$L_g^c,L_d^c=L_g,L_d$
	\State
	$r_g,r_d=|(L_g^c-L_g^p)/L_g^p|,|(L_d^c-L_d^p)/L_d^p|$
	\State
	$L_g^p,L_d^p=L_g,L_d$
    \EndWhile
  \end{algorithmic}
\end{algorithm}

%%%%%%%%% Convergence analysis
\subsection{Convergence evaluation}
In this section, we demonstrate that with our proposed update strategy, WGAN can reach local convergence for Dirac-GAN, whereas it cannot do so with a fixed update strategy.
The GAN objective function is given by:
\begin{equation}
\begin{split}
L(\theta,\psi)= & E_{ p_D(x)}[f(D_\psi(x))] +
\\ & E_{p(z)}[f(-D_\psi(G_\theta(z)))]
\label{eq: final objective function}
\end{split}
\end{equation}

\begin{figure}[th]
\centerline{
\begin{tabular}{cc}
  \resizebox{0.225\textwidth}{!}{\rotatebox{0}{
  \includegraphics{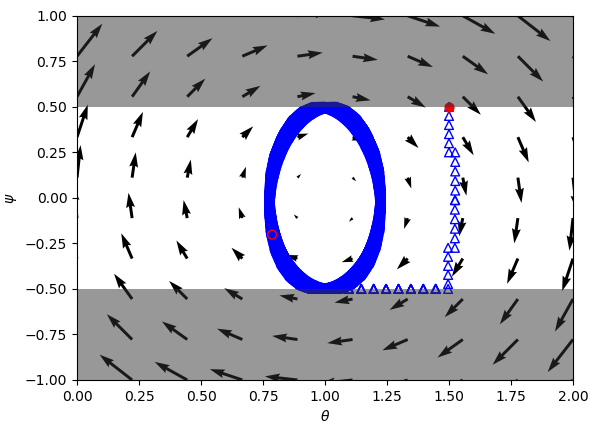}}}
  &
  \resizebox{0.225\textwidth}{!}{\rotatebox{0}{
  \includegraphics{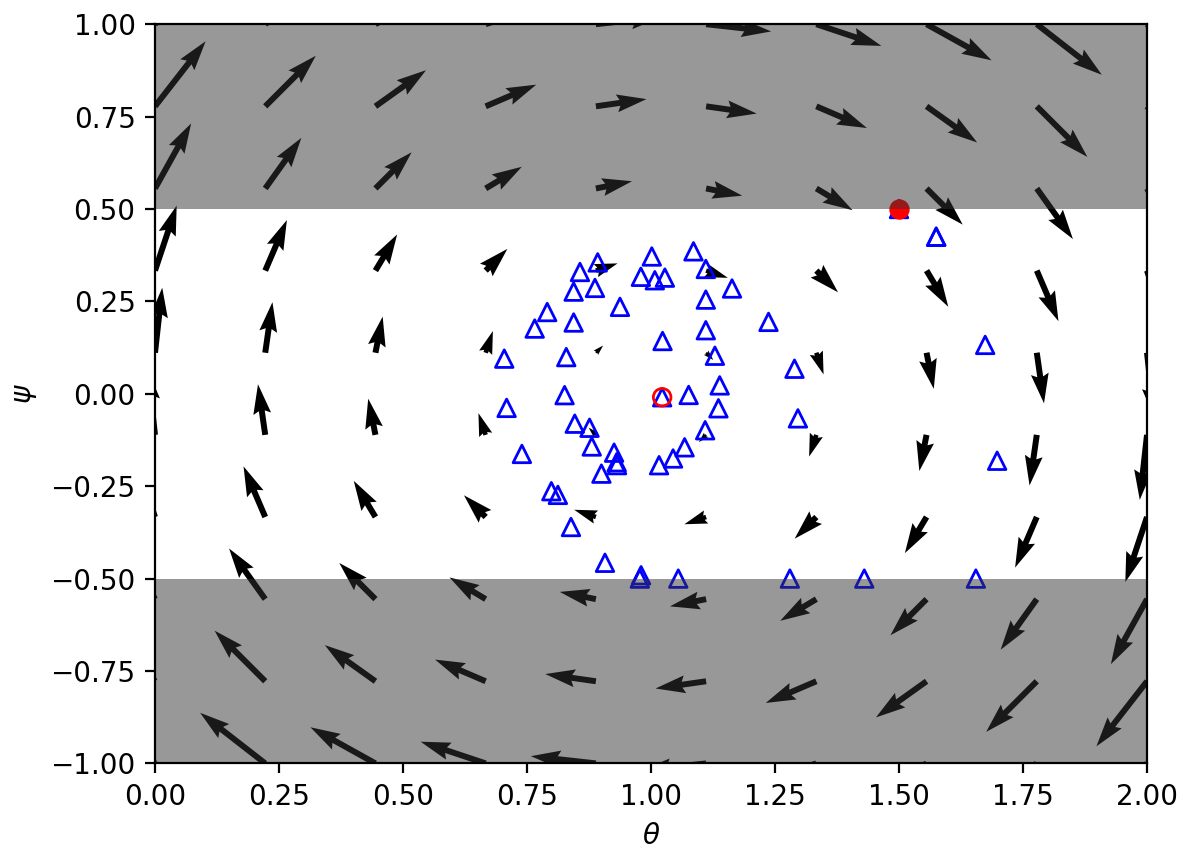}}}
  \\
  (a) fixed updates & (b) adaptive updates
\end{tabular}}
\caption{Gradient convergence of different GANs for the Dirac-GAN problem. The shadow areas in Figures (a) and (b) mark the Lipschitz constraint on discriminator parameter (-0.5,0.5). The initial point (1.5, 0.5) is marked in red and the end point is marked in hollow red. Ideally, the end point should be (1.0, 0.0). Note that the granularity in the proposed adaptive approach is not fine and so the path traced by it is somewhat less uniform.
} 
\label{fig: converge property}
\end{figure} 

\begin{table*}[bh]
\begin{center}
{\small
\begin{tabular}{|c|c|c|}
\hline
 Model & Generator loss & Discriminator loss\\
\hline
WGAN & $D(G(z))$ & $D(x)-D(G(z))$\\
\hline
WGAN-GP & $D(G(z))$ & $D(x)-D(G(z))+10*Gradient\_Penalty$\\
\hline
Deblur GAN & $D(G(z))+100*content\_loss$ & $D(x)-D(G(z))+10*Gradient\_Penalty$\\
\hline
SR WGAN & $D(G(z))+content\_loss+perceptual\_loss$ & $D(x)-D(G(z))$\\
\hline
\end{tabular}
}
\end{center}
\caption{List of WGAN loss functions group evaluated in this paper.}
\label{tab: loss function}
\end{table*}

The discriminator attempts to maximize this function whereas the generator attempts to minimize it. The goal is to find a Nash-equilibrium, where both components cannot improve their utility. 
The optimization is normally done using an alternating gradient descent where when training the generator the parameters are updated by:
\begin{equation}
\begin{split}
\theta_{t+1}&= \theta_{t}+\alpha \cdot v(\theta_{t})\\
\psi_{t+1}&=\psi_{t}
\end{split}
\label{eq: theta update}
\end{equation}
where $\alpha$ is the learning rate and $v$ is the gradient vector field; and when training the discriminator the parameters are updated by:
\begin{equation}
\begin{split}
\psi_{t+1}&= \psi_{t}+\alpha \cdot v(\psi_{t})\\
\theta_{t+1}&=\theta_{t}
\end{split}
\label{eq: psi update}
\end{equation}

The Dirac-GAN~\cite{ICML2018Mescheder} consists of a generator distribution $p_\theta = \delta_\theta$ and a linear discriminator $D_{\psi}(x)=\psi \cdot x$. The true data distribution $p_D$ is given by a Dirac-distribution concentrated at 1. Thus, there is one parameter $\theta$ in the generator and one parameter $\psi$ in the discriminator. For WGAN, we define $f(t)=t$ and add a Lipschitz constraint (-0.5, 0.5) on the discriminator as in the original WGAN. Thus, the GAN objective function in Equation~\ref{eq: final objective function} is given by:
\begin{equation}
L(\theta,\psi)=\psi \cdot 1 - \psi \cdot \theta
\label{eq: dirac gan loss}
\end{equation}

The unique equilibrium point of the objective function in Equation~\ref{eq: final objective function} is $\theta=1, \psi=0$. Since $v(\theta, \psi)=0$ if and only if $(\theta,\psi)=(1,0)$ as shown by:
\begin{equation}
v(\theta,\psi)=
\begin{pmatrix} \psi\\
1-\theta
\end{pmatrix}
\label{eq: parameter gradient}
\end{equation}
Thus, when training the generator, the parameters update in Equation~\ref{eq: theta update} are given by: 
\begin{equation}
\begin{pmatrix} \theta_{t+1}\\
\psi_{t+1}
\end{pmatrix}
=
\begin{pmatrix} 1&\alpha\\
0&1
\end{pmatrix}
\begin{pmatrix} \theta_{t}\\
\psi_{t}
\end{pmatrix}
\label{eq: generator update}
\end{equation}
When training the discriminator, the parameters update in Equation~\ref{eq: psi update} are given by:
\begin{equation}
\begin{pmatrix} \theta_{t+1}\\
\psi_{t+1}
\end{pmatrix}
=
\begin{pmatrix} 1&0\\
-\alpha+\frac{\alpha}{\theta_t}&1
\end{pmatrix}
\begin{pmatrix} \theta_{t}\\
\psi_{t}
\end{pmatrix}
\label{eq: discriminator update}
\end{equation}

We employ our proposed update strategy using an alternating gradient descent based on Equations~\ref{eq: generator update} and~\ref{eq: discriminator update}. We decide on the component to update by comparing the loss change ratios ($r_d$ and  $r_g$) as described in Algorithm~\ref{alg: WGAN with our updating strategy}. The coefficient $\lambda$ is set to 1.
For comparison we also apply the original WGAN update strategy of alternating gradient descent with fixed update steps (5 discriminator updates for each generator update). The results are shown in Figure~\ref{fig: converge property}.
As can be observed in sub-figure (a) fixed WGAN updates ($n_d=5,n_g=1$) do not converge whereas in sub-figure (b) adaptive updates following the proposed approach do converge to the Nash equilibrium point $(1,0)$.

%%%%%%%%% Network
\subsection{Network architectures}

To demonstrate our proposed adaptive training strategy as described in Algorithm~\ref{alg: WGAN with our updating strategy} we evaluated several network architectures with and without adaptive training. Specifically, we evaluated standard WGAN~\cite{Arxiv2017Arjovsky} and WGAN-GP~\cite{NIPS2017Gulrajani} networks for image synthesis, Deblur GAN~\cite{CVPR2018Kupyn} for image debluring using a Conditional Adversarial Network~\cite{Arxiv2014Mirza}, and Super Resolution WGAN~\cite{Github2018Justin} for increasing image resolution using perceptual loss, content loss, and WGAN loss. 
Except for modifying the update strategy to become adaptive, we retained the original optimizer and loss functions in each of the evaluated networks, as can be found in the papers referenced above. We show the loss function for each model as in Table~\ref{tab: loss function}.

In our update strategy we set the coefficient $\lambda$ to different values. We observe that higher values of $\lambda$ (up to 10) perform better when the generation task is complex (e.g. in Deblur GAN and Super Resolution WGAN). Smaller values of $\lambda$ ($1\sim5$) work for the WGAN and WGAN-GP image generation networks. Increasing the value of $\lambda$ results in training more the generator which is necessary due to the increased complexity of the generator. Note that training the generator more is in contrast to the suggestion in the original WGAN~\cite{Arxiv2017Arjovsky} paper where it is suggested to perform 5 training steps for the discriminator for each step of the generator.
Experimental evaluation results are provided in the next section.

%%%%%%%%% Experiments
\section{Experimental evaluation}
In this section, we train different GANs and compare with our updating strategy: WGAN, WGAN-GP, TTUR, Gradient Penalty, Deblur GAN and Super Resolution WGAN, and evaluate them both in quantitative and qualitative ways. 

\begin{figure*}[th]
\center{
\begin{tabular}{ccccc}
\hline
 Epoch & 1 & 10 & 100 & best
  \\ \hline
  \shortstack{WGAN \\ with $n_d=5, n_g=1$
  \\ \qquad \\ \qquad \\ \qquad
  \\ \qquad \\ \qquad \\ \qquad
  \\ \qquad \\ \qquad \\ \qquad}&
  \resizebox{0.15\textwidth}{!}{\rotatebox{0}{
  \includegraphics{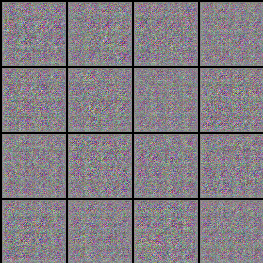}}}
  &
  \resizebox{0.15\textwidth}{!}{\rotatebox{0}{
  \includegraphics{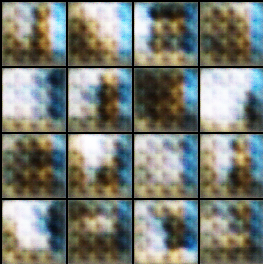}}}
    &
  \resizebox{0.15\textwidth}{!}{\rotatebox{0}{
  \includegraphics{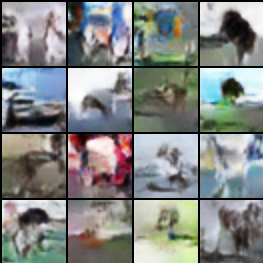}}}
    &
  \resizebox{0.15\textwidth}{!}{\rotatebox{0}{
  \includegraphics{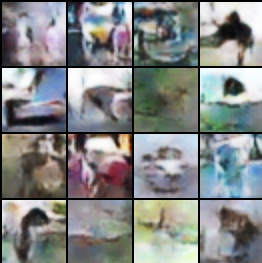}}}
  \\ \hline
   \shortstack{adaptive WGAN \\ with $\lambda=1$
   \\ \qquad \\ \qquad \\ \qquad
   \\ \qquad \\ \qquad \\ \qquad
   \\ \qquad \\ \qquad \\ \qquad}&
    \resizebox{0.15\textwidth}{!}{\rotatebox{0}{
  \includegraphics{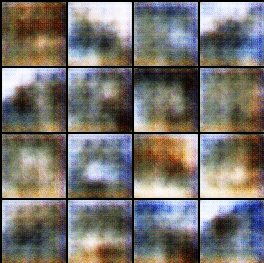}}}
  &
  \resizebox{0.15\textwidth}{!}{\rotatebox{0}{
  \includegraphics{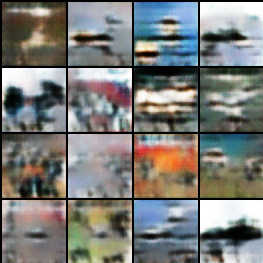}}}
    &
  \resizebox{0.15\textwidth}{!}{\rotatebox{0}{
  \includegraphics{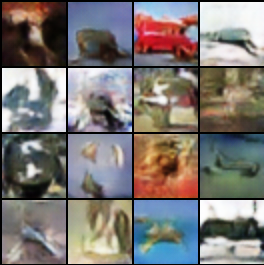}}}
    &
  \resizebox{0.15\textwidth}{!}{\rotatebox{0}{
  \includegraphics{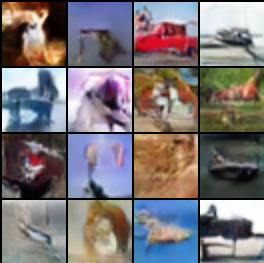}}}
   \\ \hline \hline  
    \shortstack{WGAN-GP \\ with $n_d=5, n_g=1$
    \\ \qquad \\ \qquad \\ \qquad
    \\ \qquad \\ \qquad \\ \qquad
    \\ \qquad \\ \qquad \\ \qquad}&
    \resizebox{0.15\textwidth}{!}{\rotatebox{0}{
  \includegraphics{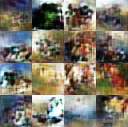}}}
  &
  \resizebox{0.15\textwidth}{!}{\rotatebox{0}{
  \includegraphics{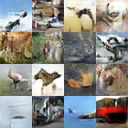}}}
    &
  \resizebox{0.15\textwidth}{!}{\rotatebox{0}{
  \includegraphics{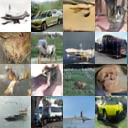}}}
    &
  \resizebox{0.15\textwidth}{!}{\rotatebox{0}{
  \includegraphics{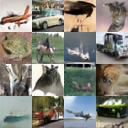}}}
    \\ \hline
  \shortstack{adaptive WGAN-GP \\ with $\lambda=1$ 
  \\ \qquad \\ \qquad \\ \qquad
  \\ \qquad \\ \qquad \\ \qquad
  \\ \qquad \\ \qquad \\ \qquad}&
    \resizebox{0.15\textwidth}{!}{\rotatebox{0}{
  \includegraphics{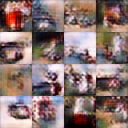}}}
  &
  \resizebox{0.15\textwidth}{!}{\rotatebox{0}{
  \includegraphics{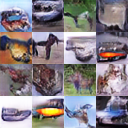}}}
    &
  \resizebox{0.15\textwidth}{!}{\rotatebox{0}{
  \includegraphics{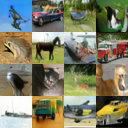}}}
    &
  \resizebox{0.15\textwidth}{!}{\rotatebox{0}{
  \includegraphics{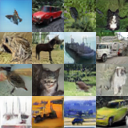}}}
    \\ \hline \\
\end{tabular}}
\caption{Examples of generated images using WGAN and the proposed adaptive WGAN trained on the CIFAR-10 dataset. The parameters $n_g$, $n_d$ are the fixed number of update steps for the generator and discriminator (as suggested in the original papers). In order to ignore the coefficient $\lambda$, we set it to 1. Results are shown for several epochs.}
\label{fig: cifar10 converge}
\end{figure*} 

\subsection{Experimental setup}
We train both WGAN and adaptive WGAN for 500 epochs and using RmsProp optimizer (learning rate=0.00005 for both G and D). We train WGAN-GP and adpative WGAN-GP for 20k iterations and using Adam~\cite{ICLR2015KingMa} optimizer (learning rate=0.0001 for both G and D, $\beta_1=0.5, \beta_2=0.9$). And both in original WGAN and WGAN-GP training, we follow the setting in~\cite{Arxiv2017Arjovsky}~\cite{NIPS2017Gulrajani} that updates five times for D per updating one time for G. In adaptive WGAN and adaptive WGAN-GP training, we tried different $\lambda$: 1,3,5,10. 

Meanwhile, we compare with the WGAN-GP TTUR (learning rate of G: $lr_g$ = 0.0001, learning rate of D: $lr_d$ = 0.0003 in TTUR; in adaptive WGAN-GP, $lr_g$=0.0003 and  $lr_d$=0.0003 in order to keep in step), Ubalanced GAN, Gradient Penalty (we follow the hyper-parameters setting in original Gradident Penalty) and with TTUR ($lr_g$=0.0001 and  $lr_d$=0.0003) and with our strategy ($lr_g$=0.0003 and  $lr_d$=0.0003 in order to keep in step).  

We train both Deblur GAN and adaptive Deblur GAN for 1500 epochs and using Adam optimizer (learning rate=0.0001 for both G and D, $\beta_1=0.5, \beta_2=0.999$). We train Deblur GAN 5 time on D and 1 time on G in each batch which followed the ~\cite{CVPR2018Kupyn}. In adaptive Deblur GAN training, we tried $\lambda$: 1, 10.

We train both Super Resolution WGAN and adaptive Super Resolution WGAN for 1000 epochs and using RmsProp optimizer (learning rate=0.001 for both G and D). We train Super Resolution WGAN one time on D and one time on G in each batch which followed the~\cite{Github2018Justin}. In adaptive Super Resolution WGAN training, we tried $\lambda$: 1, 10.

%\begin{comment}
\begin{figure*}[ht]
\centerline{
\begin{tabular}{ccc}
  \resizebox{0.24\textwidth}{!}{\rotatebox{0}{
  \includegraphics{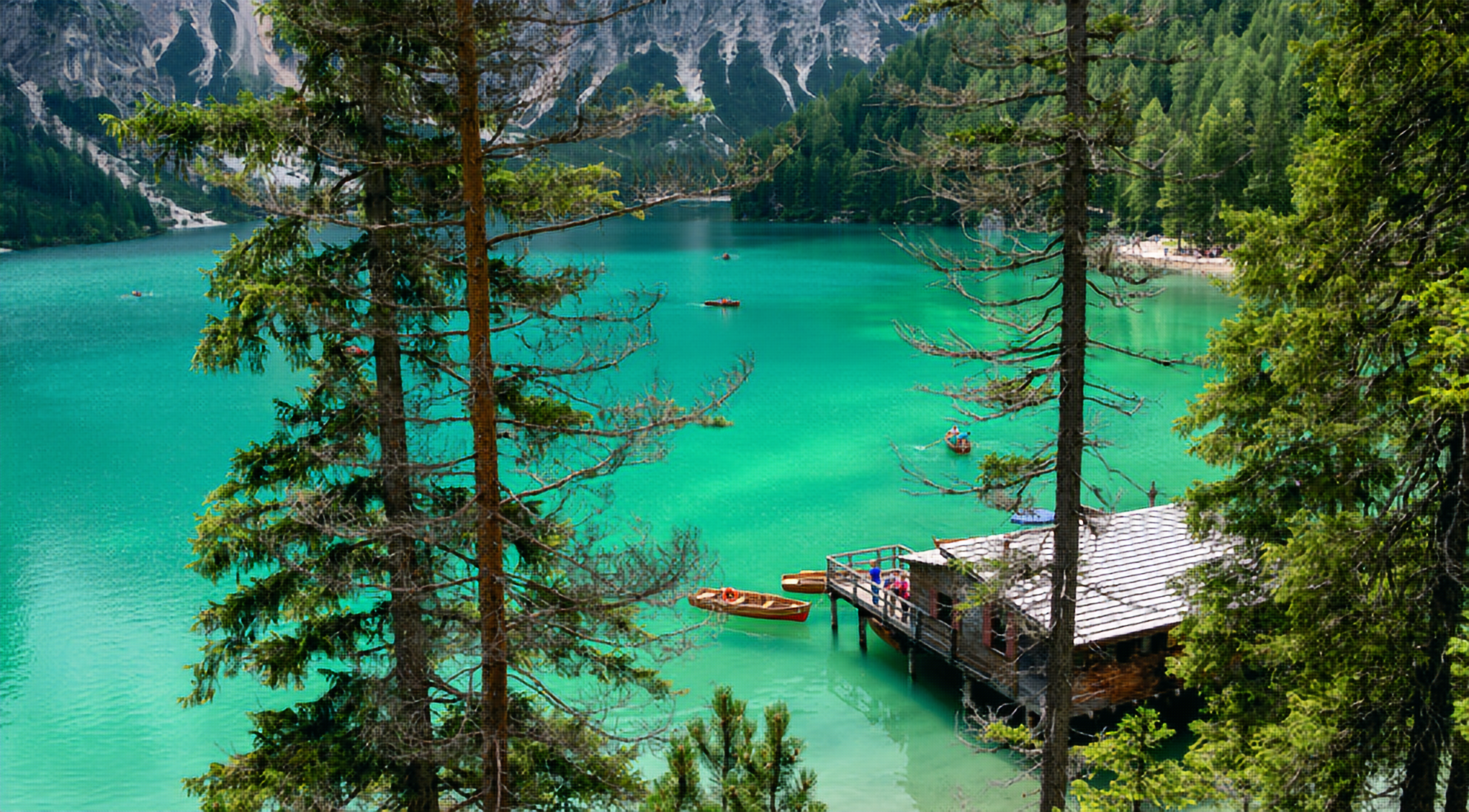}}}
  &
  \resizebox{0.24\textwidth}{!}{\rotatebox{0}{
  \includegraphics{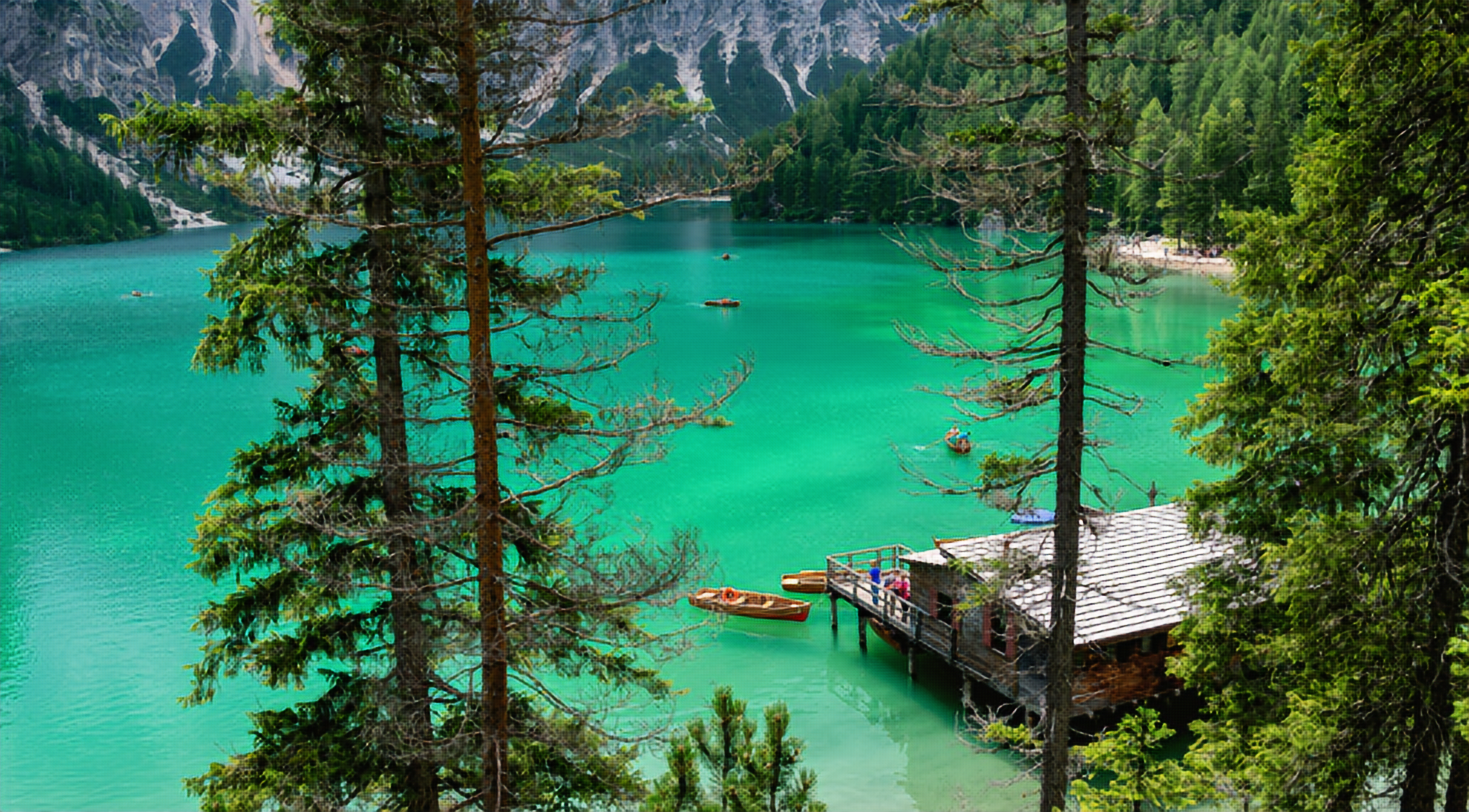}}}
  &
  \resizebox{0.24\textwidth}{!}{\rotatebox{0}{
  \includegraphics{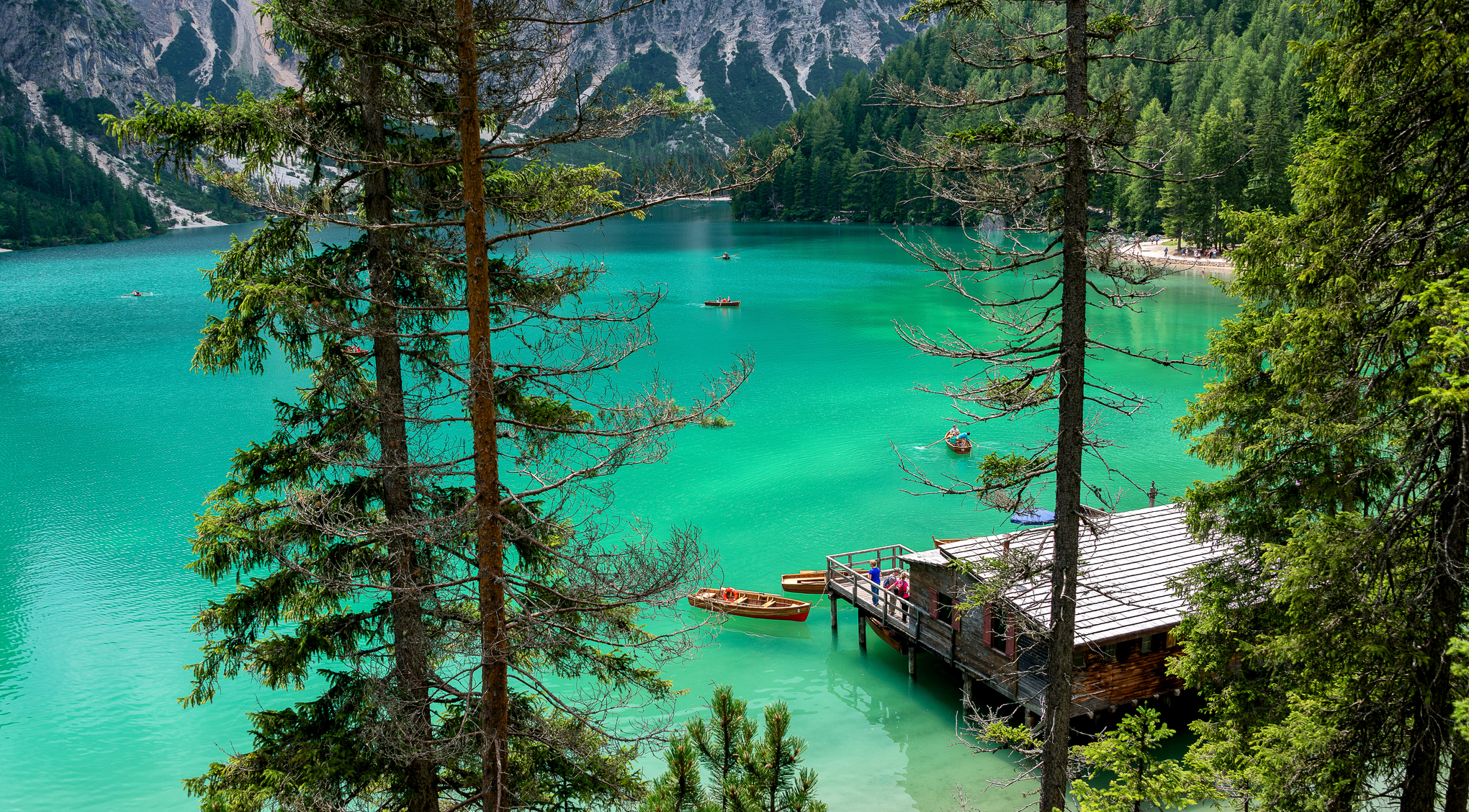}}}
  \\ 
  (a) SR-WGAN & (b) Our strategy & (c) real image\\
\end{tabular}}
\caption{Comparison of Super resolution WGAN (SR-WGAN) using a fixed and the proposed adaptive update strategy on the x4 DIV2K dataset (input image size is $510\times282$, target image size is $2040\times1182$). (a) fixed update; (b) adaptive update; (c) target image. The proposed adaptive strategy (b) gets similar results to the fixed strategy (a) with faster convergence speed.}
\label{fig: task3 visualization}
\end{figure*} 
%\end{comment}

%%%%%%%%% Datasets
\subsection{Datasets}
To train the WGAN we use a 100 dimensional random noise vectors as input. For targets we use 3 datasets: the CIFAR-10~\cite{NIPS2012Krizhevsky} which includes 50,000 training examples and 10,000 validation examples; the LSUN~\cite{Arxiv2015Yu} conference room dataset which has 229,069 training examples and 300 validation examples; and the labeled faces in the wild (LFW~\cite{LFWTech}) dataset which has 13,233 examples split into 10587 training examples, 1323 validation examples, and 1323 testing examples. For all of three datasets, we set up the image size to 64x64 to match the original paper~\cite{Arxiv2017Arjovsky} setting. 

To train the WGAN-GP and compare with the TTUR, Gradient Penalty, we all use the 128 dimensional random noise vectors as the input, and the CIFAR-10 dataset as the targets where the images are resized to 32x32 to match the original paper~\cite{NIPS2017Gulrajani} setting.

To train the Deblur GAN we use the Caltech-UCSD Birds-200-2011~\cite{WahCUB200} dataset which has 200 classes of bird images with size 256x256. We synthesize blurred images from the original images using a sequence of six 3x3 Gaussian kernel convolutions. We then use the synthetic blurred images as the inputs and the corresponding original images as targets. 

To train the Super Resolution WGAN (SR-WGAN) we use the DIV2K~\cite{CVPR2017Agustsson} dataset containing a diverse set of RGB images. In this set there are 700 training images, 100 validation images, and 100 test images. The images in this dataset are of various sizes. To synthesize the source data we downscale each image by a factor of two, 4 times thus resulting in images having 1/16 size in each spatial dimension. The original images are then used as the corresponding targets.

%\begin{comment}
\begin{figure}[ht]
\centerline{
\begin{tabular}{ccc}
  \resizebox{0.15\textwidth}{!}{\rotatebox{0}{
  \includegraphics{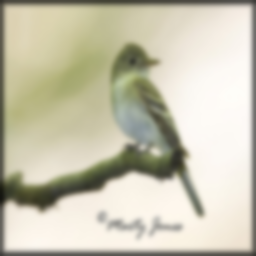}}}
  &
%    \resizebox{0.15\textwidth}{!}{\rotatebox{0}{
%  \includegraphics{pic/bird_real.png}}}
%  \\
%  (a) blurred image 
%& (b) deblurred image\\
    \resizebox{0.15\textwidth}{!}{\rotatebox{0}{
  \includegraphics{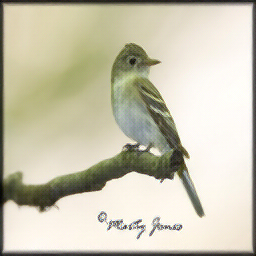}}}
   &
  \resizebox{0.15\textwidth}{!}{\rotatebox{0}{
  \includegraphics{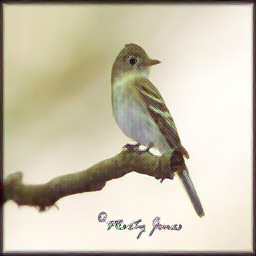}}}
    \\ 
%  (c) Deblur GAN & (d) Our Strategy\\
  (a) Blurred image & (b) Deblur GAN & (c) Our Strategy\\
\end{tabular}}
\caption{Comparison of Deblur GAN updated with a fixed and the proposed adaptive update strategy evaluated on the CUB-200-2011 dataset (image size is $256\times256$). (a) blurred image; 
%(b) target image; 
(b) fixed update; (c) adaptive update. The proposed adaptive strategy (c) gets similar results to the fixed strategy (b) with faster convergence speed.}
\label{fig: task2 visualization}
\end{figure} 

\begin{table*}[ht]
\begin{center}
{\small
\begin{tabular}{|l|l||c|c|c|c||c|c||c|c|}
\hline
\multicolumn{1}{|c|}{Model} & \multicolumn{1}{c||}{Parameters} & \multicolumn{4}{c||}{FID (reached epoch)} & \multicolumn{2}{c||}{Best epoch}& \multicolumn{2}{c|}{Total epochs}\\ 
\hline
  & & 200 & 100 & 50 & 40 & IS & FID & $u_g$ & $u_d$\\ 
 \hline \hline
 WGAN & $n_d=5$, $n_g=1$ & 17 & 46 & * & * & 3.45 & 63.62 & 65166 & 325834\\ 
  \hline
 adaptive WGAN & $\lambda=1$ & \textcolor{red}{6} & \textcolor{red}{19} & 456 & * & 3.99 & 48.10 & 253793 & 137206\\ 
  \hline
 adaptive WGAN  & $\lambda=3$ & 10 & 20 & \textcolor{red}{158} & \textcolor{red}{312} & \textcolor{red}{4.61} & \textcolor{red}{35.81} & 304873 & 86126\\ 
  \hline
 adaptive WGAN  & $\lambda=5$ & 10 & 28 & 172 & 339 & 4.54 & 37.32 & 323075 & 67924\\ 
  \hline
 adaptive WGAN  & $\lambda=10$ & 17 & 64 & 431 & * & 4.46 & 45.95 & 358568 & 32431\\ 
  \hline
\end{tabular}
}
\end{center}
\caption{Comparison of fixed WGAN update strategy with the proposed adaptive WGAN update strategy with different  coefficients $\lambda$ trained on the CIFAR10 dataset (batchsize=64, imagesize=64x64, 500 epochs). Columns 3-6 show the first epoch that reached the target FID value. Columns 7-8 show the result of the best epoch up to 500. The last to columns show the actual number of updates of G and D at the best epoch. As can be observed, while the fixed training strategy gives preference to training D, the proposed adaptive strategy ends up giving preference to training G and by doing so converges faster and to a better result. The parameters $n_g, n_d$ is the set number of fixed update steps. A $'*'$ indicates the score could not be reached by the training strategy. Note that the proposed adaptive strategy with a default parameter $\lambda=1$ performs better than a fixed update strategy and so the adaptive strategy performs better than a fixed strategy without using this parameter.
 }
\label{tab: WGAN eval}
\end{table*}

\begin{table*}[ht]
\begin{center}
{\small
\begin{tabular}{|l|l||c|c|c|c|c||c|c|c|c|c|}
\hline
  \multicolumn{2}{|c||}{} & \multicolumn{5}{c||}{LSUN Dataset} & \multicolumn{5}{c|}{LFW Dataset}\\
\hline
\multicolumn{1}{|c|}{Model} & \multicolumn{1}{c||}{Parameters} & \multicolumn{3}{c|}{FID (reached epoch)} & \multicolumn{2}{c||}{Best epoch} & \multicolumn{3}{c|}{FID (reached epoch)} & \multicolumn{2}{c|}{Best epoch}\\ 
\hline
  &  & 200 & 100 & 50 & IS & FID & 200 & 100 & 50 & IS & FID\\ 
 \hline \hline
 \shortstack{WGAN} &  $n_d=5$, $n_g=1$ & 8 & 42 & * &  3.58 & 135.41 & 75 & 211 & * & 2.60 & 55.82\\ 
  \hline
 \shortstack{adaptive WGAN} &  $\lambda=1$ & \textcolor{red}{1} & \textcolor{red}{18} & 456 & \textcolor{red}{3.72} & 133.89 & \textcolor{red}{33} & \textcolor{red}{73} & \textcolor{red}{211} & 2.58 & 38.13\\ 
  \hline
 \shortstack{adaptive WGAN} &  $\lambda=10$ & 5 & \textcolor{red}{18} & \textcolor{red}{82} & 3.61 & \textcolor{red}{114.26} & 34 & 77 & 360 & \textcolor{red}{2.64} & \textcolor{red}{37.22}\\ 
  \hline
\end{tabular}
}
\end{center}
\caption{Comparison of fixed WGAN update strategy with the proposed adaptive WGAN update strategy with different coefficients $\lambda$ trained on the LSUN conference room and LFW datasets (using the same experiment setting as in Table~\ref{tab: WGAN eval}). The proposed training strategy converges faster and to a better result.
Note that the proposed adaptive strategy with a default parameter $\lambda=1$ performs better than a fixed update strategy and so the adaptive strategy performs better than a fixed strategy without using this parameter.}
\label{tab: WGAN eval on LSUN and LFW}
\end{table*}

\begin{table*}[!hbt]
\begin{center}
{\small
\begin{tabular}{|l|l||c|c|c||c|c|}
\hline
\multicolumn{1}{|c|}{Model} & \multicolumn{1}{c||}{Parameters} & \multicolumn{3}{c||}{FID (reached epoch)} & \multicolumn{2}{c|}{Best epoch}\\
\hline
  &  & 100 & 50 & 30 & IS & FID\\ 
  \hline \hline
 \shortstack{WGAN-GP } & $n_d=5$, $n_g=1$ & 3 & 39 & * & 8.09 & 34.83\\ 
  \hline
 \shortstack{WGAN-GP TTUR} & $n_d=1$, $n_g=1$ & \textcolor{red}{2} & 32 & * & \textcolor{red}{8.41} & 33.26\\ 
 \hline 
 \shortstack{adaptive WGAN-GP} & $\lambda=1$ & \textcolor{red}{2} & 22 & 285 & 8.18 & 30.36\\ 
 \hline  
 \shortstack{Unbalanced GAN} & $n_d=1$, $n_g=1$ & / & / & / & 3.0 & /\\ 
  \hline
 \shortstack{Gradient Penalty} & $n_d=1$, $n_g=1$ & 10 & 44 & 269 & 5.35 & 27.82\\ 
  \hline
 \shortstack{Gradient Penalty TTUR} & $n_d=1$, $n_g=1$ & 8 & 26 & 130 & 5.63 & \textcolor{red}{24.14}\\ 
 \hline 
  \shortstack{adaptive Gradient Penalty }& $\lambda=1$ & 3 & \textcolor{red}{15} & \textcolor{red}{75} & 5.79 & 25.04\\ 
 \hline 
\end{tabular}
}
\end{center}
\caption{Comparison of various WGAN training schemes targeting balancing G and D training (see Section \ref{sec:related-work}). Methods with ``adaptive'' in their name employ the proposed adaptive training scheme.
%Compare the WGAN-GP, WGAN-GP with TTUR , Unbalanced GAN, Gradient Penalty, Gradient Penalty with TTUR, 
The network was trained using the CIFAR-10 dataset (batchsize=64, imagesize=32x32, 1000 epochs). The symbol '/' indicates that information is not available from the original paper. As can be observed, the proposed adaptive training scheme converges faster.
Note that the proposed adaptive strategy with a default parameter $\lambda=1$ performs better than a fixed update strategy and so the adaptive strategy performs better than a fixed strategy without using this parameter.
}
\label{tab: WGAN-GP eval}
\end{table*}

%%%%%%%%% Qualitative
\subsection{Qualitative evaluation}
Figure~\ref{fig: cifar10 converge} shows generated images with WGAN and WGAN-GP trained on the CIFAR-10 dataset usidg a fixed training strategy and the proposed adaptive training strategy. As can be observed, the proposed adaptive WGAN training strategy progresses faster than a fixed training strategy. 
When comparing the best epoch (the epoch with the best FID) results, we observe that the adaptive WGAN results are more realistic compared with the original WGAN results. Both WGAN-GP and the adaptive WGAN-GP can get some meaningful results, but the adaptive strategy progresses faster.
%% COMMENTED OUT FOR SPACE
\begin{comment}
Figure~\ref{fig: task1 visualization} shows images generated with WGAN and the proposed adaptive WGAN trained on the LSUN conference rooms and the LFW datasets. We observe that the adaptive WGAN results have more details and are more realistic on both datasets.  
\end{comment}

%\begin{comment}
Figure~\ref{fig: task3 visualization} shows example results using the Super resolution WGAN with a fixed and the proposed adaptive strategy on the DIV2K dataset. Figure~\ref{fig: task2 visualization} shows example results using the Deblur GAN  with a fixed and the proposed adaptive strategy on the synthetic CUB-200-2011 bird dataset. In both examples, the proposed adaptive strategy reaches the  results of the fixed update strategy faster. Thus, we demonstrate that we can accelerate training while maintaining stability and convergence. A quantitative evaluation is provided in the next section.
%\end{comment}

%%%%%%%%% Quantitative
\subsection{Quantitative evaluation}

To evaluate WGAN and WGAN-GP for image synthesis we use the Inception Score (IS)~\cite{IS} and the Frenchet Inception Distance (FID)~\cite{FID}. We train the networks using both the fixed strategy and the proposed adaptive strategy (with different coefficient $\lambda$ values) on the CIFAR-10 dataset and record the first epoch when a target FID value is obtained. We record in addition the total number of G and D updates. The results are shown in Table~\ref{tab: WGAN eval}.  
As can be observed the proposed adaptive training scheme converges faster and to a better result when $\lambda$ is between 1 and 5 (with best result at $\lambda=3$). Note that a higher IS score is better whereas a lower FID score is better. 

When considering the total number of G and D updates, we see that with $\lambda=3$ the proposed adaptive training strategy trains G three to four times more than D, whereas the suggested ratio in the fixed training scheme~\cite{NIPS2014Goodfellow}~\cite{ICLR2016Radford} ~\cite{Arxiv2017Arjovsky} is to train D five times more than G.
The update frequency of G and D can be effected by multiple factors such as the complexity of the models, the optimizer and its parameters (e.g. learning rate), and the loss function. It is therefore difficult to empirically estimate the G and D training ratio. Moreover, even if the training ratio of G and D is somehow determined (e.g. hyper-parameter search) it may change during iterations as the algorithm gets close to convergence. The proposed adaptive training strategy alleviates the need to carefully set this parameter and provides a systematic way to continuously estimate it. While the proposed adaptive scheme still involves selecting an coefficient $\lambda$, training results are less sensitive to the selection of this parameter and a simple default (e.g. $\lambda=1$) may suffice.

\begin{table*}[!hbt]
\begin{center}
{\small
\begin{tabular}{|l|l||c|c|c||c|c||c|c|}
\hline
\multicolumn{1}{|c|}{Model} & \multicolumn{1}{c||}{Parameters} & \multicolumn{3}{c||}{PSNR (reached epoch)} & \multicolumn{2}{c||}{Best epoch}& \multicolumn{2}{c||}{Total epochs}\\ \hline
  & & 24 & 25 & 25.5 & PSNR & SSIM & $u_g$ & $u_d$\\ 
 \hline \hline
 Deblur GAN & $n_d=5$, $n_g=1$ & 30 & 126 & 1452 & 25.6323 & 0.7350 & 784833 & 3924167\\ 
  \hline
 adaptive Deblur GAN & $\lambda=1$ & 8 & 31 & \textcolor{red}{38} & 25.6461 & 0.7427 & 3195559 & 1513441\\ 
  \hline
 adaptive Deblur GAN & $\lambda=10$ & \textcolor{red}{5} & \textcolor{red}{16} & 97 & \textcolor{red}{25.7806} & \textcolor{red}{0.7500} & 4608989 & 100011\\ 
  \hline
\end{tabular}
}
\end{center}
\caption{Comparison of Deblur GAN trained without and with the proposed adaptive training scheme. The network was trained using the CUB-200-2011 bird dataset (image size=256x256). The coefficient $\lambda$ in the proposed adaptive update strategy is attempted with different values. Columns 3-5 show the first epoch that reached the target PSNR value. The last two columns show the total number of training epochs for D and G. As can be observed the proposed adaptive training scheme trains the generator more than the discriminator whereas when employing the suggested fixed strategy trains the discriminator more than the generator. We observe that the proposed adaptive training strategy (with all $\lambda$ values) converges faster and get a better result.
}
\label{tab: Deblur eval}
\end{table*}

\begin{table*}[!hbt]
\begin{center}
{\small
\begin{tabular}{|l|l||c|c|c||c|c||c|c|}
\hline
\multicolumn{1}{|c|}{Model} & \multicolumn{1}{c||}{Parameters} & \multicolumn{3}{c||}{PSNR (reached epoch)} & \multicolumn{2}{c||}{Best epoch}& \multicolumn{2}{c|}{Total epochs}\\ \hline
 & & 25 & 26 & 27 & PSNR & SSIM & $u_g$ & $u_d$\\ 
 \hline \hline
 Super resolution WGAN  & $n_d=1$, $n_g=1$ & 40 & 82 & 546 & \textcolor{red}{26.8646} & 0.7630 & 43837 & 43837\\ 
  \hline
 adaptive Super resolution WGAN  & $\lambda=1$ & 30 & 102 & 395 & 26.6840 & 0.7555 & 37899 & 49775\\ 
  \hline
 adaptive Super resolution WGAN  & $\lambda=10$ & 22 & 65 & 224 & 26.8230 & \textcolor{red}{0.7677} & 64592 & 23082\\ 
  \hline
\end{tabular}
}
\end{center}
\caption{Comparison of the super resolution WGAN (SR-WGAN) trained without and with the proposed adaptive training scheme. The network was trained using the DIV2K dataset (image size=2040x1182). The coefficient $\lambda$ in the proposed adaptive update strategy is attempted with different values. Columns 3-5 show the first epoch that reached the target PSNR value. The last two columns show the total number of training epochs for D and G. We observe that the proposed adaptive training strategy (with $\lambda=10$) converges faster and get a better result.
}
\label{tab: Super eval}
\end{table*}

A similar evaluation of the proposed adaptive WGAN training strategy when trained with different datasets is provided in Table~\ref{tab: WGAN eval on LSUN and LFW}. In this table training is done separately both with the LSUN conference room and LFW datasets. The evaluation on these datasets results in similar  conclusions to the ones obtained when training with the CIFAR-10 dataset. The proposed adaptive training strategy converges faster and to a better result.

Comparison of various WGAN training schemes targeting balancing G and D training (see Section 2) is provided in Table~\ref{tab: WGAN-GP eval}. The compared methods include: WGAN-GP~\cite{NIPS2017Gulrajani}, WGAN-GP with TTUR~\cite{NIPS2017Martin}, Unbalanced GAN~\cite{Arxiv2020Ham}, and Gradient Penalty~\cite{ICML2018Mescheder}. Training is done using the CIFAR-10 dataset.
Methods with ``adaptive'' in their name employ the proposed adaptive training scheme.
As can be observed, the proposed adaptive training scheme converges faster.

%Gradient Penalty got the best results in IS since they have deeper neural network with condition label inputs (~\cite{ICML2018Mescheder} mentioned WGAN-GP got the similar final inception scores to Gradient Penalty when they have the same structure). TTUR WGAN-GP claimed get the lowest FID, however, they did not give the neural network details in the paper. Since we only implement our strategy on the WGAN-GP, and as we can see, the adaptive strategy still can accelerate the convergence speed, and the results are better than or approximate to the original WGAN-GP.

%% tables 5-6

The deblur GAN and super resolution WGAN networks are evaluated in Tables~\ref{tab: Deblur eval} and~\ref{tab: Super eval} respectively. The Deblur GAN network is trained using the 
Caltech-UCSD Birds-200-2011 dataset whereas the SR-WGAN network is trained on the DIV2K dataset. 
Training in both cases is done both with a fixed update strategy and the proposed adaptive training strategy. Evaluation is done using SSIM~\cite{SSIM} and PSNR where in both a higher score means better results. In the evaluation we record the first epoch during training where a target PSNR value is achieved. In addition we record the total number of update steps for the generator and the discriminator. 
We observe that the proposed adaptive training scheme converges faster and to a better result for both the deblur and super resolution networks. Further, here too the ratio of training steps for G and D obtained by the proposed adaptive training strategy does not match the recommended ratio of training steps thus supporting the need for the proposed adaptive training scheme.

%In Table~\ref{tab: Deblur eval}, we can see that when $\lambda=10$, the adaptive strategy can reach the approximate evaluation results to the Deblur GAN with nearly six times converge speed acceleration. Besides, when $\lambda=10$, $u_g:u_d\approx46:1$, and when $\lambda=100$ the strategy prefers to update most time on G rather than D in first 120 epochs (since PSNR reaches 25.5 at 119 epoch which is quite close to the best results in original Deblur GAN). This situation also happens in Table~\ref{tab: Super eval}: when $\lambda=10$, we get five times convergence speed accelerating compared to Super resolution WGAN reach PSNR=27 and the model trains the G almost all the time.  Combined with the results from WGAN training evaluation, we could conclude that with our adaptive updating strategy, the neural network with WGAN loss can reach the optimal goal faster and save the redundant time on updating discriminator. Furthermore, the controllable penalty coefficient $\lambda$ is do helpful to accelerate the convergence speed.

\section{Conclusion}
In this paper, we propose an adaptive WGAN training strategy which automatically determines the sequence of generator and discriminator training steps. The proposed approach compares the loss change ratio of the generator and discriminator to decide on the next component (G or D) to be trained and so balances the training rate of the generator and discriminator.
We show that a WGAN with this strategy could reach the local Nash Equilibrium point for the Dirac-GAN. 
Experimental evaluation results using different networks and datasets show that the proposed adaptive training scheme normally converges faster and to a lower minimum. Another advantage of the proposed adaptive update strategy is that it alleviates the need to empirically determine the number of update steps for the generator and discriminator.
In future work, we will investigate additional update strategies suitable for various GAN structures and loss terms.

%------------------------------------------------------------------------
%\newpage
%\clearpage
{\small
\bibliographystyle{ieee_fullname}
\bibliography{egbib}
}

\end{document}